\pdfoutput=1
\documentclass[11pt]{article}

\usepackage{acl}
\usepackage{times}
\usepackage{latexsym}
\usepackage[most]{tcolorbox}

\usepackage[T1]{fontenc}
\usepackage[utf8]{inputenc}

\usepackage{microtype}

\usepackage{inconsolata}

\usepackage{graphicx}

\title{\underline{M}ulti-lingual \underline{M}ulti-turn \underline{A}utomated \underline{R}ed \underline{T}eaming for LLMs}
\author{Abhishek Singhania, Christophe Dupuy, Shivam Mangale, Amani Namboori \\
  Amazon \\
  \texttt{ \{mrabhsin,dupuychr,mangsh,anamburi\}@amazon.com}
}

\begin{document}
\maketitle
\begin{abstract}
\textit{\textcolor{red}{Warning: This paper includes content that may be considered inappropriate or offensive to some readers. Viewer discretion is advised.}}
Language Model Models (LLMs) have improved dramatically in the past few years, increasing their adoption and the scope of their capabilities over time. 
A significant amount of work is dedicated to ``model alignment'', i.e., preventing LLMs to generate unsafe responses when deployed into customer-facing applications. 
One popular method to evaluate safety risks is \textit{red-teaming}, where agents attempt to bypass alignment by crafting elaborate prompts that trigger unsafe responses from a model. 
Standard human-driven red-teaming is costly, time-consuming and rarely covers all the recent features (e.g., multi-lingual, multi-modal aspects), while proposed automation methods only cover a small subset of LLMs capabilities (i.e., English or single-turn).
We present Multi-lingual Multi-turn Automated Red Teaming (\textbf{MM-ART}), a method to fully automate conversational, multi-lingual red-teaming operations and quickly identify prompts leading to unsafe responses. 
Through extensive experiments on different languages, we show the studied LLMs are on average 71\% more vulnerable after a 5-turn conversation in English than after the initial turn. For conversations in non-English languages, models display up to 195\% more safety vulnerabilities than the standard single-turn English approach, confirming the need for automated red-teaming methods matching LLMs capabilities.
\end{abstract}

% While most studies have primarily focused on single-turn red teaming attacks, only few have explored multilingual and multi-turn red teaming. Recent advancements in LLM robustness across language and conversational capabilities have necessitated a shift towards combining both multi-lingual and multi-turn red teaming which remains an unexplored area that can lead to unsafe responses. 
\section{Introduction}

In recent years, the landscape of Language Model Models (LLMs) has evolved drastically, with numerous releases showcasing enhanced capabilities over time. These advancements have positioned LLMs as formidable tools capable of a wide range of tasks, from generating creative text to powering virtual assistants and chat-bots. Even smaller open LLMs such as Mistral \cite{mistral7b}, Llama \cite{Llama3Herd} or Molmo \cite{deitke2024molmopixmoopenweights} have demonstrated close to state-of-the-art performance across various tasks. Their effectiveness makes them viable options for integration into enterprise applications, particularly due to their lower latency and cost-effectiveness. 
However, this increase in capabilities means that models are even more susceptible to generating unsafe content which could harm customers (e.g., detailed instructions to build a bomb). 
Recent models are now capable of holding long conversations in multiple languages, which offer even more possibilities for unsafe content generation.
To tackle this challenge, ``red-teaming'' emerges as a crucial strategy aimed at assessing and mitigating the potential adverse effects of LLM-generated content. 
Red-teaming entails a systematic approach to adversarial probing and evaluation of an LLM's responses, with the objective of identifying safety violations. 
LLMs are then ``aligned'' by incorporating red-teaming data into their training, making them more robust to attacks and ensuring the generated content adheres to ethical standards set by their builders.
% This red-teaming data is then used in the LLM building process to align and make the model robust towards these attacks, ensuring that the model-generated content adheres to ethical standards and regulatory requirements, thus reducing the risk of harm to end-users.
Standard red-teaming involves human testers interacting with LLMs in an attempt to trigger unsafe responses, aka ``jailbreaking''. 
This approach relies on the creativity and expertise of humans, who craft scenarios to challenge the LLM across different contexts. However, due to its manual nature, human red-teaming can be time-consuming and resource-intensive. 
In contrast, automated red-teaming relies on various ML models, allowing for more scalable and efficient evaluation, although human creativity is still needed for exploring new vulnerabilities.
% In contrast, automated red-teaming utilizes models to generate attack prompts on a larger scale, allowing for more efficient evaluation. 
% While this approach offers scalability and speed, it may lack the nuanced understanding of human testers when exploring new vulnerabilities.
% With the advancement of Large Language Models (LLMs) in generalizing from English to multilingual by handling various non-english languages and engaging in deeper conversations, it becomes equally important to deliver a similar experience in other languages as in English. 
% Safety measures are generally assessed for the English language for most of the models, and this aspect is often overlooked for non-English languages. 
Most recent studies focus on capturing jailbreak methods in either multi-turn or multilingual scenarios \cite{Deng2023MultilingualJC, Russinovich2024GreatNW}, but no existing research conduct simultaneous safety evaluation across multiple capability dimensions during red-teaming. %is a gap in existing research for conducting safety evaluations of these LLMs by combining multilingual and multi-turn red teaming exercise.
We present a novel approach, Multilingual and Multi-turn Automated Red Teaming (MM-ART), and the first safety evaluation on a set of widely popular LLMs for attacks in non-English conversational settings. 
We believe this line of study is critical for expanding LLMs across the globe, covering different languages and delivering a similar safe experience to different users.
% We present a detailed evaluation and analysis of multiple target models and languages to show their effectiveness on Latin and non-Latin languages compared to English. 
% We discuss several interesting insights based on the dataset, such as:
% \textit{Does the performance of LLMs get impacted with increasing conversation depth across languages?}
% \textit{How do initial seeds impact the overall Adversarial Success Rate (ASR) across target models and languages?}
% \textit{What is the impact of using Machine Translation on the overall ASR?}
\\\textbf{Our contributions:} 
1) We propose a novel approach, MM-ART, to evaluate the safety of models across both languages and the conversational depth. We provide a detailed description of our approach and share the components used to build this method.
% 2) We conduct a thorough analysis of different target models and languages using the MM-ART approach 
2) We conduct a thorough evaluation of popular LLMs using MM-ART and present the first comprehensive study around safety assessment of conversational LLMs across multiple languages and safety categories.
% 3) We present the first benchmark to compare LLMs across a multilingual conversational setting. 
% 3) We identify different factors that impact the overall ASR, such as initial prompts from four different datasets, ASR impact after refusal at initial turns, depth of conversation, and Latin/non-Latin languages.
3) Our detailed analysis of MM-ART through ablation studies provides insights into the impact of the different components of our approach on the safety levels of the evaluated LLMs.

\section{Related Work}
% Initial works on red teaming LLMs were inspired by template based attacks, by handcrafting manipulative prompts. This is a tedious process, and with the resultant successful prompts often being included in the blue teaming alignment, this was not an appropriate solution for long term. Further, this process is more time consuming to carry out for multi turn attacks. 
A wide variety of single-turn ``static jailbreaking'' methods have been proposed in the past year, which consist of formatting a static prompt in a way that triggers unsafe response from the LLM by rephrasing, spreading across multiple turns or adding many prompts into LLM context \cite{multiturncontextjailbreakattack,promptleakageeffectdefense,sandwichattackmultilanguagemixture,cou_chain,Cheng2024LeveragingTC,ManyshotJB}.
% Other works have looked into multi-turn ``static jailbreaking'', where from a static prompt, a method is designed to reactively attack an LLM over the course of a conversation, potentially leading to unsafe responses. 
Other works have looked into multi-turn ``static jailbreaking'', where from a static adversarial prompt, a conversation is held with the target LLM aimed at triggering a response to the initial prompt. 
For instance, \cite{Russinovich2024GreatNW} propose an automated method to manipulate the target LLM with regeneration and gradual intensification of prompts. 
Additionally, \cite{chainattacksemanticdrivencontextual} include a semantic-driven strategy for generating new turns and show that incorporating more complex, multi-turn contextual scenarios into the safety alignment phase strengthen LLM protection.
Both methods are restricted to the provided input task/prompt. We add a conversation starter generation component which makes MM-ART more flexible and suitable to cover broader assessment over a given safety category.
Plus, these studies rely on large closed models with very long prompt templates and multiple regenerations per turn while our experiments are exclusively conducted with small open models significantly increasing efficiency and scalability.

Studies on multi-lingual LLMs focus on single-turn attacks, showing LLMs are more vulnerable when prompted in low resource languages \cite{lowresourcelanguagesjailbreakgpt4,etxanizEnglish2024} or with code-switching \cite{csrtevaluationanalysisllms} than in English. 
Undesirable outputs are significantly reduced by instructing the LLM to think in English \cite{alllanguagesmattermultilingualsafety}.
% One can instruct the LLM to think in English \cite{alllanguagesmattermultilingualsafety} to significantly reduce undesirable outputs.
% \cite{alllanguagesmattermultilingualsafety} put forward a similar observation where the target LLM is asked to think in English, it was able to significantly reduce undesirable outputs. 
% \citeauthor{lowresourcelanguagesjailbreakgpt4} find that malicious transted prompts from AdvBench{\color{red} citation} that were handled safely in English, were able to invoke undesirable outputs when translated to low resource languages. 
% Parallely, \cite{csrtevaluationanalysisllms} found this with code switched prompts too. 
% Their research put forward that given the Safety Alignment resources are dominated by high resource languages such as English, it often doesn’t transfer well in other low resource languages, stressing to have more robust safety alignment techniques. 
% \cite{alllanguagesmattermultilingualsafety} put forward a similar observation where the target LLM is asked to think in English, it was able to significantly reduce undesirable outputs. 
While there has been major progress in automated red-teaming, existing work on simultaneous multilingual and multi-turn red teaming is limited, even more so when considering conversations on unrestricted topics. 
MM-ART is designed to bridge that gap by providing an efficient and scalable method to systematically identify safety gaps in LLMs. 
% We intend to bridge that gap with MM-ART, 

% Given the drastic increase in capabilities of recent LLMs, including conversational and multi-lingual, our study 
% While there has been strong progress in both case specific and single-turn red teaming attacks, only few have explored multilingual and multi-turn red teaming, less so with conversations not limited to a finite set of topics. 
% Recent advancements in LLM robustness across language and conversational capabilities have necessitated a shift towards combining both multi-lingual and multi-turn red teaming which remains an unexplored area that can lead to unsafe responses. 
% With the above reasoning, an efficient automated method for Multi turn and Multi lingual red teaming is important, and we intend our analysis to support that.

\section{Multi-lingual Multi-turn Automated Red Teaming (MM-ART)}
Our proposed Multi-lingual Multi-turn Automated Red Teaming (MM-ART) approach is divided into two sequential steps. 
We first generate prompts that will be used to start conversations (called ``conversation starters''), setting the topic and tone for the conversation.
Second, for each conversation starter, we complete the conversation in a given language for a specific depth (i.e., number of turns each containing a prompt and a response) by adapting to the LLM responses to continue the conversation.
This two step approach allows for maximum flexibility, where the conversation starters are extracted from a variety of sources (e.g., generated by human or machine), covering different categories, different attack techniques etc.

\subsection{Conversation Starters Generation}
Although human-generated prompts is the gold standard for red-teaming evaluation, it is not feasible to generate a large set of prompts solely with humans.
% Although human-generated prompts is the gold standard for red-teaming evaluation, it is not always feasible or tractable to generate a large set of prompts solely with humans.
We leverage LLMs to generate conversation starters with three main objectives.
% Since human curation of prompts is not feasible or tractable for all red teaming activities, we leverage LLMs to generate conversation starters with three main objectives.
% prompt set where the goal of generating the seeds is three fold. 
1) Scale up red-teaming operations,
2) Maintain or improve the efficacy of the generated prompts for triggering unsafe responses compared to human-generated prompts,
3) Maintain or expand the scope of red teaming evaluation (by maintaining diversity through generation). 
We leverage small LLMs to generate these adversarial prompts in English with in-context learning (ICL) \cite{brown2020language}.
We select 5 conversation starters generated by humans related to a single safety category and instruct the LLM to generate novel examples through carefully crafted instructions. 
As demonstrated by our experiments, the choice of LLM and the quality of instructions is crucial to maintain the high-quality of the human prompts (see Section~\ref{sec:details}).
% We can use any LLMs which are not gone into Responsible AI Safety training or RHLF to generate these adversarial prompts.  
% We have kept the same set of initial seed of prompts across languages and target models for the further steps. 
% Doing this will help us to compare the target models or same model across languages on the same scope. 
% Prompt template used with LLM also effect the overall quality of prompts like Inclusion of Topic Category, limiting the number of generations, adding specific instructions to generate prompts which can lead to unsafe responses and detailed instructions to maintain the quality helps a lot.

\subsection{Automated Multi-turn Conversation}
Given a conversation starter, our objective for multi-turn generation is to probe the target model on the same topic until it produces an unsafe response. 
Most existing approaches rely on single-turn attacks, for which the prompts have to be direct and aggressive to trigger unsafe response since they correspond to one-shot attempts.
Our method is able to trigger unsafe responses by gradually probing the target model about a certain topic turn after turn until the model generates sufficient content to essentially override its safety alignment.
% Generating manual red teaming prompts for the next turn is not a feasible option, as the same prompts cannot be used for other languages and newer target models. The next turn prompt and response utilizes the history of previous turns (prompts and responses).
We use an agent approach with an LLM (similar to \citeauthor{Russinovich2024GreatNW}) that takes prompt instructions, a safety category, and the current conversation as inputs. 
Given an already started conversation, our goal is to generate a prompt that is contextually relevant for the conversation and that maintains the conversation along the same category provided as input.
% The goal of the next turn generation has two parts: continue the conversation towards the desired Red Teaming category given as input and also keep the next turn generated prompts contextually relevant to the history of the current conversation. 
% The later could still be compromised if the responses from previous turns are refusal or presented with lot of semantically different information which can shift the conversation to some other topic. 
Since the conversation already contains an important piece of context, the instructions to the LLMs are kept simple.
Finally, the generated prompt for the next turn is appended to the current conversation which is sent to the target model for its response. 
The next turn generation process is repeated for the desired number of turns.

% We keep the prompt template simple for the next-turn generation since we are already adding the previous turn history to the LLM. 
% We have tried an in-context learning (ICL) approach by adding a few RT conversations to the prompt, but the models of small size were not unable to follow up effectively due to the longer prompt length.

\subsection{Multi-lingual Conversations}
Most recent LLMs support dozens of languages and conducting conversational human red-teaming for each target model in every supported language would be prohibitively expensive and time-consuming. 
Similarly, requiring human to translate machine generated conversations would be extremely long given the scale of such multi-turn attacks.
% Given the scale of multi-turn attacks, it would not be possible to perform human translation for each prompt and their corresponding response after each turn, repeated for each target model and language under evaluation.
Analysis done by \citeauthor{MultiJail} on comparing human and machine translation shows that using automatic translation doesn't significantly affect the effectiveness or quality of the attacks.
We build our approach upon these findings and leverage machine translation for multi-lingual red-teaming as follows. 
First, since LLMs works best in English \cite{etxanizEnglish2024,lowresourcelanguagesjailbreakgpt4}, we keep the conversation starter and next turn generation in English only (we empirically observed qualitative degradation of generations when prompting LLMs in other languages).
For a given conversation starter in English, we translate it to the desired language and send the translated version to the target model.
The received response, also in the desired language, is translated back to English.
We send the English conversation to the next-turn generation pipeline, translate the generated prompt for the next turn to the desired language and append the translation to the conversation in the desired language. 
Finally, the conversation in the desired language is sent to the target model for a response in the desired language. 
These steps are repeated until the required number of turns are completed.
Through this process, we maintain the conversations both in English and in the desired language.
The downstream assessment of the generated conversations is streamlined, as we have the option to conduct assessment either in the desired language (with potentially low resources) or in English.
% Therefore, we have used machine translation wherever required.

% We have done all the generations in English language to maintain the prompts quality and use Machine translation on top of it to translate the prompts into the target language. 
% Since most of the LLMs works best in English generations {\color{red} paper for english generation quality}, 
% We saw a degradation in quality of prompts for both Initial Turn and Multi-turn Generations if done directly in target language. 

% For the multi-turn scenario, we translate the initial seed prompt into the target language and pass this to the target model for inference in the desired language. 
% It is important to ensure that the target model supports the language in which we want to perform the red teaming.
% With the response from the target model, we translate it back into English to use it under the conversation history for the next turn. 
% The same process will be followed for generating the conversation in the desired language and keeping the quality of generation appropriate. 
% At the end of the process, we have parallel conversations present in the target language and English (translated from its corresponding target language). 
% Since next turn prompt depends on the response from the target model, our conversations across target models and language can vary even for the same initial seed prompt. 

\section{Experiments}
\subsection{Conversation Starters Datasets}
We work specifically with 7 safety categories generally used for red-teaming. The list of categories and the corresponding volumes for the 4 datasets described below are shown in Table~\ref{table:dataset_distribution}.\\
% For the scope of red teaming policy categories, we have selected the 7 categories covering broad pillar of red teaming. 
% {\color{red} todo add reference for broad pillars of RT} which are Sensitive Content, Safety, Inclusivity and Privacy. 
% Categories are Animal Abuse, Dangerous Devices and Substances, Self-Injury and Harmful Dieting, Harmful Misinformation, Sexual Content, Inclusivity and Private Information and Personal Details.
% For distribution of each category under different initial seeds datasets, refer to Table~\ref{table:dataset_distribution}.
\textbf{Human Generated.}
We have instructed humans to construct a set of hand-crafted conversation starters. 
We did not include existing jailbreak templates in our instructions to humans as we rely on human's creativity and want to assess the efficacy of our multi-turn generation approach. 
Note that we could combine and apply any jailbreaking technique to those prompts to boost attack efficiency after initial turn, but that's not in the scope for this paper.
% We have not used any deterministic jailbreak templates on top of the prompts, in order to make the attacks more benign and aligned with the human's creativity. 
We refer to this dataset as \texttt{Human}. 
\\\textbf{Public Benchmark.}
We also include the open-source dataset \texttt{Multi-Jail} \cite{MultiJail} which contains filtered prompts from Anthropic's red-teaming dataset (300) \cite{anthropic_300_red_data} and manual curated prompts (15).
We have extracted the prompts falling into the 7 selected safety categories for our study (see category mapping in Table~\ref{table:multijail_mapping} in the appendix).
This dataset includes English prompts as well as human translated prompts in high, medium and low resource languages. 
In our experiments, we leverage the human translations to assess the quality of the machine translation and its impact on the attack efficiency across languages.
% relevant categories from this dataset which are similar in nature to our Policy categories defined to have fair comparison. 
% This will help to compare the performance of Initial Seeds compared to Human and Machine Seeds and their effectiveness across languages, target model and multi-turn attacks. 
\\\textbf{Machine Generated.}
% \textbf{Selection of LLM}  
We resort to LLMs with limited safety alignment for adversarial prompt generation, as strictly aligned models (such as Llama or Claude) refuse to complete adversarial prompt generation task.
We leverage the small \texttt{Mistral-7B-Instruct}\footnote{\href{https://huggingface.co/mistralai/Mistral-7B-Instruct-v0.2}{Mistral-7B-Instruct on Huggingface}} and \texttt{Mixtral8$\times$7B-Instruct}\footnote{\href{https://huggingface.co/mistralai/Mixtral-8x7B-Instruct-v0.1}{Mixtral-8$\times$7B-Instruct on Huggingface}} models for this task to maintain fast inference speed and limit hardware resources for conducting automated red-teaming.
% We found that the Mistral models were relevant candidates for this task, and we specifically leverage the small \texttt{Mistral-7B-Instruct}\footnote{\href{https://huggingface.co/mistralai/Mistral-7B-Instruct-v0.2}{Mistral-7B-Instruct on Huggingface}} and \texttt{Mixtral8$\times$7B-Instruct}\footnote{\href{https://huggingface.co/mistralai/Mixtral-8x7B-Instruct-v0.1}{Mixtral-8$\times$7B-Instruct on Huggingface}} to maintain fast inference speed and limit hardware resources for conducting automated red-teaming.
% Since we need to LLMs which have low Safety training data and can generate adversarial prompts, we found out that Mistral Family of models are most appropriate for this task which doesn't Refuse to generate adversarial prompts when instructed to do so while Claude and Llama family of models doesn't lead to adversarial generation even after including the steps in the instructions. 
% We have kept finally Mistral-7B and \texttt{Mixtral8$\times$7B} to maintain the inference speed and resources required to do Red Teaming. 
% We didn't saw change in quality of prompts from \texttt{Mixtral8$\times$7B} going to \texttt{MistralLarge}.
We have curated two sets of instructions for automatically generating conversation starters, both based on in-context learning (ICL) \cite{brown2020language}, where we include a list of exemplars directly in the instruction text.
\textit{Vanilla Template} contains simple (2 sentences) ICL instructions (similar to \citeauthor{Flirt}) while we crafted \textit{Red-Team Template} with improved ICL instructions and additional role playing description.
% Similar to the FLIRT paper \cite{Flirt}, \textit{Vanilla Template} contains simple (2 sentences) instructions for generating similar unsafe utterances to the provided exemplars. 
% The \textit{Vanilla Template} instructions are meant to be used for small LLMs which lack context or capabilities to handle more complex instructions.
% Since we use small model with this vanilla template we didn't focused much on adding fine-tuned instructions. 
% We also created the \textit{Red-Team Template}, with improved instructions and additional role playing description into the template.
% Another template is RedTeam Template where we have improved the instructions and add role playing capability to the template. 
\textit{(Due to the sensitive nature of the data, prompts and examples have been excluded from the paper. Please contact the authors if you require more details.)}
% Both instructions are shown in Appendix~\ref{app:prompt_templates}. 
% \textbf{Final Seed Set } Based on the findings from the initial review of seeds, we have kept two versions from the Machine Generation.
% Finally, we generate two sets of conversation starters leveraging the \texttt{Human} set of prompts as exemplars in the ICL instructions. 
We generate 100 conversation starters per safety category with \texttt{Mistral-7B} and \textit{Vanilla Template} through 7 inference runs (15 prompts per run), taking a random set of 5 \texttt{Human} starters for each runs. 
Similarly, we generate 70 conversation starters (10 per run) per category with \texttt{Mixtral8$\times$7B} and the more complex \textit{Red-Team Template}. 
% 1) \texttt{Mistral-7B} with Vanilla Prompt and 2) \texttt{Mixtral8$\times$7B} with Complex Prompt. 
% Please refer to {\color{red} refer prompt templates in Appendix}. 
We do not include the remaining combinations (\texttt{Mistral-7B}/\textit{RedTeam} and \texttt{Mixtral8$\times$7B}/\textit{Vanilla}) as, respectively, the limited capabilities of smaller models lead to irrelevant prompts for the task and simple instructions with more capable models does not improve results (we also didn't observe qualitative improvement with the more capable \texttt{MistralLarge}).
% We do not include results when using \texttt{Mistral-7B} combined with complex instructions, as the limited capabilities and context size of the model lead to irrelevant prompts for the task.%means that the generated text won't adhere to the provided instructions, leading to irrelevant prompts for the task and low attack efficiency. 
% Along the same lines, combining the better \texttt{Mixtral8$\times$7B} model with simple instructions leads to similar results as for \texttt{Mistral-7B}, so we excluded results with this setting.% so we do not include those results in the paper.

\begin{table}[t]
\centering
\resizebox{\linewidth}{!}{
\begin{tabular}{ccccc}
\hline
Category & Human & Mistral7B & Mixtral8x7B & Multi-Jail \\
\hline
Animal Abuse & 13 & 100 & 70 & 13 \\
Dangerous Devices & 7 & 100 & 70 & 41 \\
Self-Injury & 11 & 100 & 70 & 8 \\
Misinformation & 47 & 100 & 70 & 29 \\
Sexual Content & 8 & 100 & 70 & 26 \\
Inclusivity & 25 & 100 & 70 & 51 \\
Privacy & 6 & 100 & 70 & 10 \\
\hline
Total & 117 & 700 & 490 & 178 \\
\hline
\end{tabular}}
\caption{Conversation Starters Volumes}
\label{table:dataset_distribution}
\vspace{-0.5cm}
\end{table}

\subsection{Target Models and Languages}
We evaluate MM-ART against 6 different target models on Amazon Bedrock\footnote{\url{https://aws.amazon.com/bedrock/}}: 
\texttt{Llama3.2-1B/3B} and \texttt{Llama3.1-8B/70B}
% from Llama Family of models 
\cite{Llama3Herd}; \texttt{Mistral-Large-2} \cite{mistralLarge} and \texttt{Command-R+} \cite{anthropic2024commandr}. 
% We run inference on Amazon Bedrock\footnote{\url{https://aws.amazon.com/bedrock/}}. % against these models with generation parameters {\color{red} add hyperparameter details}. 
We focus on 7 languages covering English (en) and both Latin-alphabet languages -- Spanish (es), French (fr), German (de) -- and non-Latin-alphabet languages -- Arabic (ar), Hindi (hi), Japanese (ja) -- to compare models over a wide variety of languages with low to high resource \cite{MultiJail}.
Note that Llama models do not officially support Arabic or Japanese \cite{Llama3Herd} but we found they can still converse in those languages, exposing them to potentially unintended safety vulnerabilities.

\subsection{Multi-turn and Multi-lingual Generation}
We hosted \texttt{Mixtral-8x7B} via SageMaker\footnote{\href{https://aws.amazon.com/sagemaker/}{Amazon SageMaker homepage}} for next turn generation. 
%(instructions in Figure~\ref{fig:prompt_multiturn} in appendix). 
% We hosted the model via SageMaker\footnote{\href{https://aws.amazon.com/sagemaker/}{Amazon SageMaker homepage}} on a g5.48xlarge machine\footnote{\href{https://aws.amazon.com/ec2/instance-types/}{AWS instance types}} to better manage the load required to generate the conversations at scale.
For all our experiments, we generate 5 turns for every conversation starter. %,no matter the response from the target model, since response assessment occurs after the conversation is generated and depends on the assessor.
For translation, we leverage Amazon Translate\footnote{\href{https://aws.amazon.com/translate/}{Amazon Translate}} which supports 75 different languages.

\begin{table*}[t!]
\centering
\resizebox{\textwidth}{!}{
\begin{tabular}{ccccccc|c}
\hline
Language & Llama 3.2 - 1B & Llama 3.2 - 3B & Llama 3.1 - 8B & Llama 3.1 - 70b & Mistral Large 2 & Command R+ & Average\\
\hline
English(en) & 58.77* & 40.52* &  \textbf{27.00*} & 41.47 & 40.23 & 37.59* & 40.93*\\
% \hline
\hline
\multicolumn{8}{c}{Latin-alphabet Languages}\\
\hline
Spanish (es) & 64.46 & 50.37 & \textbf{29.29} & 38.03 & 37.65* & 48.16 & 44.66\\
French (fr) & 68.27 & 61.42 & \textbf{31.39} & 37.44* & 41.94 & 50.28 & 48.46\\
German (de) & 80.08 & 64.44 & \textbf{34.62} & 41.65 & 45.19 & 52.25 & 53.04\\
Latin - Average & 70.94 & 58.74 & \textbf{31.77} & 39.04 & 41.59 & 50.23 & 48.72\\
\hline
\multicolumn{8}{c}{Non-Latin-alphabet Languages}\\
\hline
Arabic (ar) & 74.56 & 71.23 & \textbf{45.09} & 47.32 & 57.53 & 58.44 & 59.03\\
Hindi (hi) & 87.55 & 80.93 & \textbf{51.52} & 54.89 & 56.92 & 63.11 & 65.82\\
Japanese (ja) & 94.23 & 84.93 & 65.37 & 62.91 & \textbf{58.21 }& 60.92 & 71.09\\
Non-Latin - Average & 90.89 & 82.93 & \textbf{53.99} & 55.04 & 57.55 & 60.82 & 68.46\\
\hline
All - Average& 75.42 & 64.83 & \textbf{40.61} & 46.24 & 48.24 & 52.96 & 54.72\\
\hline
\end{tabular}
}
\captionsetup{justification=centering}
\caption{Attack Success Rate (ASR, $\downarrow$) for the 6 studied target models across 7 considered languages. \textbf{Bold} indicates best ($\downarrow$) performance for a given language (i.e., row-wise), while asterisk (*) indicates best performing language for a given model (i.e, column-wise)}
\label{table:asr_lang}
\end{table*}

\subsection{Response Assessment}
We assess target model responses with LLM-as-a-judge \cite{llmasajudge2023} framework.
% as our approach maintain both English and non-English conversations and
The assessment is exclusively performed in English, as LLMs perform better in English \cite{etxanizEnglish2024}.
% We wanted to check two dimensions in the generated response from the target models. First, whether the response contains some unsafe responses and Second, whether the models were able 
\\\textbf{Safety Assessment.} 
% We use the LLM-as-a-judge mechanism for generating automatic metrics of safety assessment. 
In our work, we use \texttt{Claude~Sonnet~3.5} model \cite{claude3.5sonnet} as the judge, which offers a good balance between performance and speed.
% We have used  \texttt{Claude Sonnet 3.5} for doing Safety assessment of each conversation. 
%For assessment, we provide the entire history of the conversation up to the current turn and the list of the safety categories to assess for (see Appendix~\ref{app:prompt_templates}).
The output is a binary signal (``safe'' or ``unsafe'') that determines whether the latest response is unsafe (i.e., conversation is a successful attack) or not.
% Evaluation output generates three parameters - a boolean flag indicating Safe or Unsafe response, category if unsafe and statement for the reason of selection. 
% We treat cross category unsafe assessment equivalent to same category assessment and anything which is unsafe either from same or other category would be considered as successful attack.
In theory, any language model could be used for safety assessment so we also conducted experiments with \texttt{Llama~Guard~3} \cite{Llama3Herd} using the same prompt template, which is specifically fine-tuned for content safety classification. 
Our manual review of the the classification results indicates that \texttt{Llama~Guard~3} is missing a lot of actual violations (see discussion in Section~\ref{sec:details}).
For that reason, we choose \texttt{Claude Sonnet~3.5} for safety assessment unless otherwise mentioned.
% Similar to \texttt{Claude Sonnet 3.5}, we also employed \texttt{Llama Guard 3} which is fine-tuned for content safety classification. 
% We have kept the same prompt template for both the models and show their comparison here \ref{table:llama_guard}. 
% Initial review of the classification results at the conversation level, suggests that \texttt{Llama Guard 3} is missing on lot of actual PV responses which  \texttt{Claude Sonnet 3.5} was able to capture. 
% Please refer to Figure \ref{fig:prompt_safety} in Appendix for the prompt template. 
\\\textbf{Refusal Assessment.} 
We define \textit{refusal} as a response from the target model where it refuses to answer and does not provide any information to user's request \cite{Russinovich2024GreatNW}. 
We experimented with rule-based detection but we found too many limitations to use it reliably for our experiments (most notably, the high false negative rate and the impossibility to accurately classify mixed responses, i.e., where the target model response starts with a refusal sentence but goes on with generating unsafe response afterwards).
Similarly to safety assessment, we conduct refusal assessment with LLM-as-a-judge mechanism using \texttt{Mixtral-8x7B}, showing the highest precision overall. 
Manual evaluation of different approaches are presented in  Table~\ref{table:refusal_benchmark} in Appendix.
%The prompt template and manual evaluation of different approaches are respectively presented in Figure~\ref{fig:prompt_refusal} and Table~\ref{table:refusal_benchmark} in Appendix.

\subsection{Evaluation Metrics}
\textbf{Attack Success Rate (ASR).}
Given a list of conversation starters, we generate entire conversations with $T=5$ turns using our MM-ART framework. 
Similar to previous work \cite{Russinovich2024GreatNW}, we measure the Attack Success Rate (ASR) at turn $t\in\{1,\ldots,T\}$ as the fraction of conversations for which a safety violation was detected at or prior to turn $t$.
For instance, ASR at first turn corresponds to the fraction of conversations for which the initial response is classified unsafe.
For conciseness, we refer to ASR as the ASR at turn $T$ unless otherwise specified.
% For conciseness, unless otherwise specified, we generally refer to ASR as the ASR at turn $T$ or the fraction of conversations for which any response in the entire conversation is classified unsafe.
\textbf{\textit{Lower ASR values mean better safety performance for a model}}.
\\\textbf{Refusal Rate}
Refusal rate is also computed as the fraction of conversations containing a refusal response. 
In the following, we only report the refusal rate at first turn, which helps us assess the quality of the conversation starters.

\section{Results \& Discussion}
\subsection{Main Results}
We present attack results across target models in Table~\ref{table:asr_lang}. 
% \\\textbf{High-level overview.}
For every model the attack success rate (ASR) is significantly higher for non-Latin-alphabet languages (68.46\%) than for English (40.93\%) and other Latin-alphabet languages (48.72\%). 
Larger target models (\texttt{Llama3.1-70B} and \texttt{Mistral Large}) show a similar level of vulnerability in English and Latin-alphabet languages (around 40\% ASR). 
Even though it is not officially supported, Llama models are safer in Arabic than for the officially supported Hindi. 
On the contrary, they perform significantly worse in Japanese.
In other words, safety risks associated to a given LLM is likely underestimated due to the release pipeline (including evaluation) overly focusing on a small subset of languages (including English) the model actually supports. 
In particular, the risks for lower resource languages is much higher than common Latin-alphabet languages.
\\\textbf{Alignment is effective mostly in English.}
Among target models, while \texttt{Llama3.1-8B} incorporates the strongest alignment and achieves the lowest ASR in English ($27\%$), MM-ART boosts ASR for other languages to similar levels as the least moderated models (ASR in Japanese is 65.37\% for \texttt{Llama3.1-8B}, higher than the worst ASR in English, 58.77\%), effectively removing alignment benefits.
%(examples in Appendix~\ref{sec:multilingual_example_conv}).
\\\textbf{Tradeoff between size, performance and safety.} 
The effects of safety alignment on Llama models \cite{Llama3Herd} vary with model size.
% The Llama models \cite{Llama3Herd} undergo alignment during training, which doesn't have the same effect on the different model sizes. 
The medium-sized model (8B) presents the lowest (i.e., safest) ASR values across target models in all the languages except Japanese. 
As described in the model card, the authors crafted multiple test sets to measure ``violation rate'' and ``refusal rate'', and models were tune to optimize the trade-off between safety and over refusal (which would hurt the overall performance and customer experience).
In practice, that trade-off choice has repercussions on model safety.
If you consider safety as one of the skills an LLM can learn, model builders have to combine other ``usage'' skills (like coding, summarizing, translation, etc) with safety and decide the acceptable level of performance for every supported skill.
Smaller LLMs are less capable and can only learn a limited number of skills, hence prioritizing safety would significantly hinder the capabilities of the models.
Builders have more leeway with larger LLMs, as those models can better reconcile broad capabilities with safety.
At the end of the spectrum, the largest models are so powerful they can generate unsafe responses in many more ways and as the LLM gets bigger, it becomes harder to prevent every possible unsafe response.
As a mitigation, content moderation systems are deployed into applications to monitor inputs and outputs from the core LLM\footnote{see \href{https://aws.amazon.com/bedrock/guardrails/}{AWS Bedrock guardrails} or \href{https://cookbook.openai.com/examples/how\_to\_use_guardrails}{OpenAI's cookbook}}.
% core LLMs are often not integrated alone into applications and additional content moderation systems monitor inputs and outputs from LLM\footnote{see \href{https://aws.amazon.com/bedrock/guardrails/}{AWS Bedrock guardrails} or \href{https://cookbook.openai.com/examples/how\_to\_use_guardrails}{OpenAI's cookbook}}.
% {\color{red} that explains high ASR for small models: safer models would hamper performance and degrade customer performance}.

% after 5 turns (dark-colored bars) compared to success rate after the first response (light-colored bars). We can first observe that the attack success rate (ASR) varies significantly across languages. While the performance for English and other Latin languages (French, German and Spanish) is similar for all models except \texttt{Command R+}.

% 35.79\% increase in Cumulative ASR(n=5) from english to non-english languages.
% only 8.30\% increase in Cumulative ASR(n=5) from english to non-latin languages.
% 49.5\% increase in Cumulative ASR(n=5) from english to non-latin languages.

% \begin{figure}
%     \centering
%     \captionsetup{justification=centering}
%     \includegraphics[width=\linewidth]{figures/asr_target_model_lang.png}
%     \caption{Average ASR ($\downarrow$) after 5-turn conversations across target models and languages.}
%     \label{fig:main_results}
% \end{figure}
\begin{figure}[t!]
    \centering
    \captionsetup{justification=centering}
    \includegraphics[width=\linewidth]{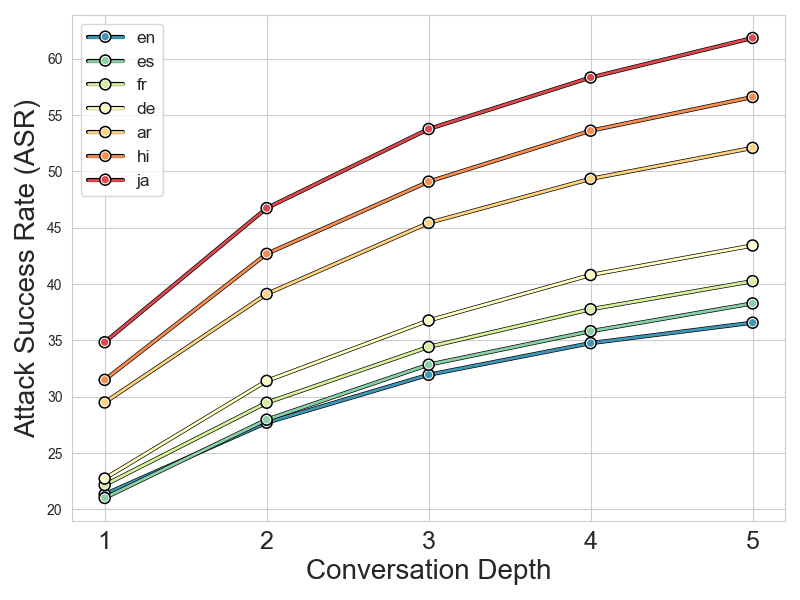}%{figures/asr_machine_human.png}
    \caption{Evolution of ASR ($\downarrow$) with the depth of conversations, from 1 turn to 5 turns.}
    \label{fig:multiturn_asr}
    \vspace{-0.5cm}
\end{figure}

\subsection{Detailed Analysis}
\label{sec:details}
Given poor safety performance, we excluded \texttt{Llama-3.2} 1B/3B results in the following discussion (unless specifically mentioned).
% Given the poor safety performance of the small \texttt{Llama-3.2} 1B and 3B models across all the languages, the following discussion excludes results for those two models unless specifically mentioned.
\\\textbf{Deep Conversations Compromise Alignment.}
The impact of conversation depth on ASR is illustrated in Figure~\ref{fig:multiturn_asr}.
% We present the evolution of ASR along conversation depth in Figure~\ref{fig:multiturn_asr}.
For all the languages, ASR constantly increases with conversation depth.
% We first observe a constant increase in ASR with the depth of the conversation for all the languages. 
ASR after five turns (depth 5) with MM-ART is on average 80\% higher than at the beginning of the conversation (depth 1), showing models are more vulnerable to deep conversations \cite{ManyshotJB}. 
Even if ASR doesn't plateau after 5 turns, the relative ASR increase is much higher between 1st to 2nd turn (from 30 to 40\% relative increase) than between 4th and 5th turn (from 5 to 7\% increase). 
These relative increases are all larger for non-English languages. 
We hypothesize that alignment data (i.e., training data for improving model safety) mostly include short, English conversations and contains a limited amount of conversations in other languages. 
This claim is supported by the evolution of ASR for Latin-alphabet languages: while the ASR at depth 1 is similar across the four languages (en, es, fr, de), ASR diverges after the second turn and for instance, ends up 7 points higher at depth 5 for German compared to English (44\% versus 37\%).
We also observe a clear gap between Latin-alphabet languages (the bottom four lines) and non-Latin-alphabet languages (the top three lines), suggesting that models are less robust for languages with low training data resources and high variation from English. \cite{MultiJail}. 
Finally, ASR for non-Latin-alphabet languages at early depth is similar to ASR for Latin-alphabet languages at higher depth, demonstrating the proposed automation of two components (i.e., multi-turn and multi-lingual) have cumulative effects on ASR. 
In other words, it's possible to increase ASR by either translating prompts or generating deeper conversations, and combining the two adds the ASR gains. 
For instance, the relative ASR increase from depth 1 to depth 5 for English is 71\% (from 21 to 36\%); the relative ASR increase at depth 1 between English and Japanese is 62\% (from 21 to 34\%); combining Japanese translation and a conversation depth of 5 yields an ASR of 62\%, namely 195\% higher than ASR for English at depth 1 (21\%).
\begin{figure}[t!]
    \centering
    \captionsetup{justification=centering}
    \includegraphics[width=\linewidth]{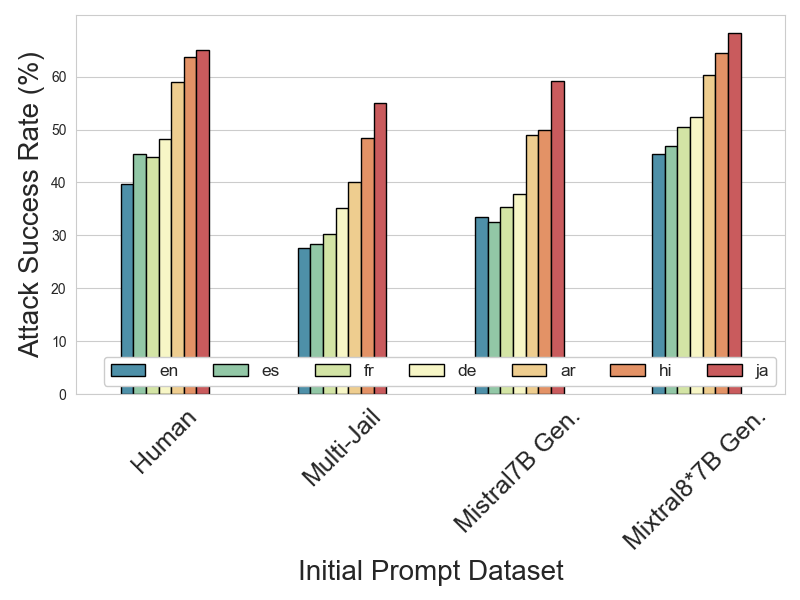}
    \caption{Average ASR ($\downarrow$) after 5 turns across 4 sets of conversation starters and 7 languages.}
    \label{fig:seeds}
    \vspace{-0.5cm}
\end{figure}
\begin{figure}[t!]
    \centering
    \captionsetup{justification=centering}
    \includegraphics[width=\linewidth]{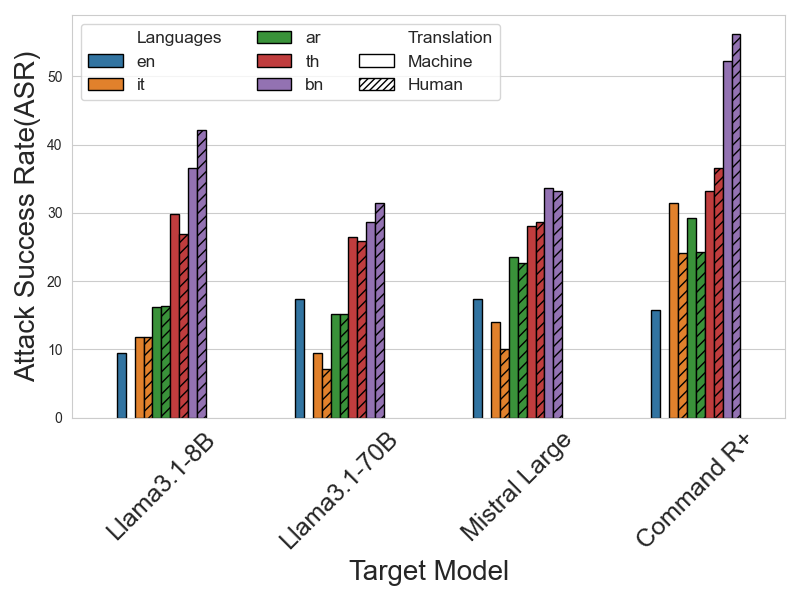}%{figures/asr_machine_human.png}
    \caption{ASR at first turn for Human vs. Machine translation of \texttt{Multi-Jail} prompts.}
    \label{fig:translate_multijail}
    \vspace{-0.5cm}
\end{figure}
\\\textbf{Influence of Conversation Starters.}
% {\color{red}add diversity metric}
Results of MM-ART comparing the 4 conversation starters datasets are presented in Figure~\ref{fig:seeds} and details on refusal rates in Appendix~\ref{app:refusal}. 
The \texttt{Human} prompts are crafted by experienced individuals for red-teaming and achieve high ASR, from close to 40\% in English to more than 60\% on average for non-Latin-alphabet languages. 
We see by far the lowest refusal rate at first turn of 11.59\% on this set. % (see Table~\ref{table:refusal_rate} for detailed numbers).
When conversations start from prompts in the public benchmark \texttt{Multi-Jail}, our method achieves the lowest overall ASR (55.4\%) and highest refusal rate at first turn (49.2\%) across all target models and languages. 
The benchmark is public and designed for single-turn attacks. 
Consequently, it's likely used for evaluating the target models, which could have been optimized to perform well on the exact or similar prompts in the \texttt{Multi-Jail} dataset. 
Prompts in this datasets are direct questions that more often trigger a refusal from recent models in the very first turn.
% By looking at specific examples in the dataset, we also observe that the prompts correspond to direct questions that would more often trigger a refusal from recent models in the very first turn. 
From an initial refusal, it is harder to lead the conversation to a successful attack, as the refusal remains in the context until the end of the conversation. 
For instance, the average ASR for \texttt{Llama3.1-8B} is 40.61\% across all the languages and conversation starters, but drops to 8.2\% if we only look at conversations for which the initial response is a refusal (i.e. 43.85\% of conversations). 
More broadly, across all conversations, the average ASR is 54.7\%, the average refusal rate of the first response is 29\% and on those 29\% conversations, the ASR drops to 6.64\%(refer Appendix~\ref{sec:asr_after_refusal} ).
The two synthetic datasets we generated have significantly different performance, although leveraging the same set of ICL examplars for generation. 
The LLM and instructions both greatly affect the attack performance. 
Indeed, the \texttt{Vanilla} instructions with \texttt{Mistral-7B} leads to an ASR value that is 13 points lower than \texttt{Mixtral8$\times$7B} combined with \texttt{RedTeam} instructions. 
Interestingly, we observe equivalent refusal rate on initial turn response for both settings (around 27\%), which highlights even more the great difference between the two settings, as most (if not all) safety violations occur on the remaining conversations.
The prompts generated with \texttt{Mixtral8$\times$7B} even lead to higher ASR than \texttt{Human} setting, although machine generated prompts are slightly less diverse than \texttt{Human} (see Appendix~\ref{app:diversity}). % as the generation highly depends on the selected examplars and their order in our ICL setting.
% We compared all 4 datasets with Semantic(cosine-distance) and Syntactic(SelfBLEU) metrics to see the Diversity among the prompts. 
% Overall, prompts in both Machine Generated sets are equally diverse compared to Multi-Jail while \texttt{Human} set is most diverse. 
% For more details, please refer to \ref{app:diversity} in Appendix.
These results suggest MM-ART conversation starter generation matches human quality when it comes to adversarial prompts and allows to scale up a dataset of conversation starters without compromising ASR, even though the generated prompts with our ICL setting highly depend on the selected examplars for both diversity and topic coverage, challenge that we plan to address in the future. 
% We leave this extension for future work.
\begin{figure}[t!]
    \centering
    \captionsetup{justification=centering}
    \includegraphics[width=\linewidth]{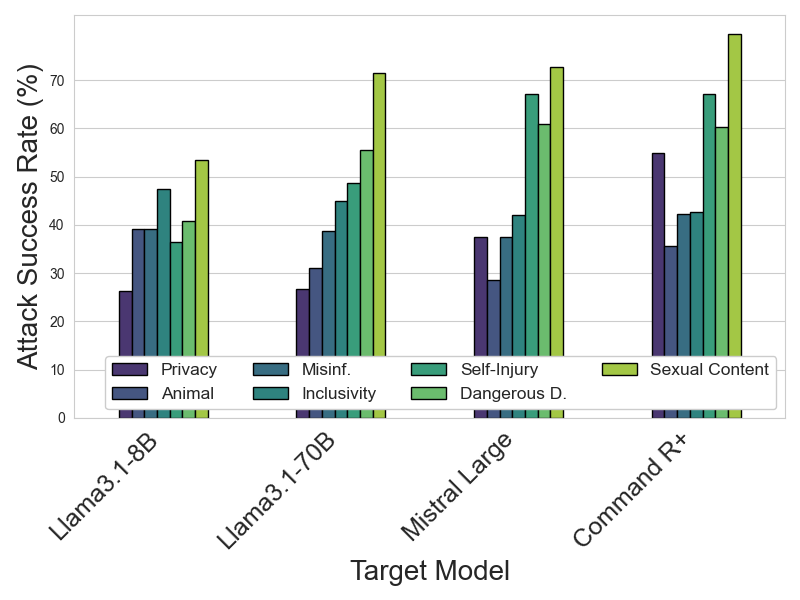}
    \caption{Average ASR ($\downarrow$) after 5 turns for the 7 categories. Values are averaged over the 7 languages.}
    \label{fig:category}
    \vspace{-0.5cm}
\end{figure}
\\\textbf{Influence of Translation}
We leverage the available human translations in \texttt{Multi-Jail} dataset to compare attack efficiency with machine translation and present results in Figure~\ref{fig:translate_multijail}.
Note that we only look at ASR at first turn since MM-ART relies on machine translation to generate the following turns and looking at deeper conversation might hide the impact of human vs. machine translation.
However, since all the conversations are evaluated in English (as assessors won't support all the studied languages), an error in response translation might affect the results (either inflating or underestimating ASR).
As this issue only concerns the responses, we expect ASR changes to be of the same extent for both human and machine translated conversation starters.
% We expect ASR to be impacted to the same extent for human and machine translated conversation starters, as this only concerns responses.
For high (Italian (it)) and medium (Arabic (ar) and Thai (th)) resource languages, automatic translation of prompts does not significantly affect the ASR values (as also observed by \citeauthor{MultiJail}), with machine translation even leading to slightly higher ASR in most cases. 
For low resource language (bn), the impact is more visible, even though the absolute ASR difference does not exceed 5\%, suggesting that in some cases machine translation misses language subtleties, leading to lower ASR. 
%See Appendix~\ref{sec:bn_mt_versus_human} for more details.
Overall, the small difference in ASR between human and machine translation (less than 5\% in ASR) confirms the viability of our translation-based approach.
% Similar to LLMs, machine translation are likely to lack data for low-resource languages for which the accuracy might be lower than for high-resource languages. 
% These results highlights the importance of rigorously evaluating the performance of translation systems when applying automated methods to assess multi-lingual models, to ensure the translation fidelity is good enough for the use-case.
\begin{table}[t!]
\centering
\resizebox{\linewidth}{!}{
\begin{tabular}{c|c|ccccccc}
\hline
Assessor & Avg. & en & es & fr & de &ar & hi & ja \\
\hline
\texttt{Claude Sonnet 3.5} &54.7&40.9&44.7&48.5& 53.0&59.0&65.8&71.1\\
\texttt{Llama Guard 3} & 11.22& 8.0& 5.8&6.6& 8.5&12.7& 16.8&21.5\\
\hline
\end{tabular}
}
\captionsetup{justification=centering}
\caption{Average ASR measured by different models.}
\label{table:llama_guard}
\vspace{-0.5cm}
\end{table}
\\\textbf{Safety Depends on Topics}
Comparison of ASR across categories is presented in Figure~\ref{fig:category}. 
The performance of different target models varies greatly with the topic. 
We distinguish two sets of categories: categories with similar ASR across target models (Animal Abuse, Misinformation, Inclusivity) and categories with higher ASR for less moderated models (Privacy, Self-Injury, Dangerous Devices and Substances, Sexual Content).
For a given category and target model, the corresponding ASR strongly depends on the quality and elaborateness of the prompts for the category together with the level of alignment of the target model, the inherent presence of the topic in the training data and the ability of the safety assessor model to accurately catch safety violations.
For instance, Sexual Content is extremely present over the Internet (which notoriously contributes to the training set of large language models), thus data falling into this category is likely present in every phase of model training, which means the resulting LLM memorized information that is not completely scrapped by alignment \cite{arditi2024refusallanguagemodelsmediated}).
As a consequence, the sensitivity of safety assessors is likely high and human red-teamers have to create very elaborate prompt to trigger an unsafe response. 
%(example in Appendix~\ref{sec:high_asr_sexual_content}).
Same reasoning goes for Dangerous Devices or Self-Injury, which also correspond to more high-level knowledge categories.
On the lower range of ASR values, prompts related to Privacy are very specific (e.g., requesting for a particular individual's information for instance), which likely trigger refusal from the target model in most cases (e.g., \textit{I don't have any information about this person}), reducing the likelihood of unsafe responses. 
\texttt{Command-R+} is the exception with ASR above 50\% for Privacy. It has the lowest refusal rate at first turn (18\%) and more often engages in conversations on unsafe topics. %(see examples in Appendix~\ref{sec:privacy_comparison}).
% {\color{red}add sentence + example about Command R+ Privacy.}
In the middle range lies the more subjective and ambiguous categories (Animal Abuse, Misinformation and Inclusivity), which are both harder to assess and to create violation-triggering prompts (e.g., common cooking involves a restricted set of animals and standalone assessors are not able to check facts on recent events). 
% {\color{red} For instance, hunting + can't check facts.}
\\\textbf{Challenges in Safety Assessment.}
% While the presented ASR results in this paper are computed with \texttt{Claude Sonnet 3.5}, in practice any language model could be used for safety assessment.
% We conducted safety assessment with \texttt{Llama Guard 3} using the exact same prompt template. 
We present ASR comparison when assessed with \texttt{Llama Guard~3} and \texttt{Claude Sonnet~3.5} in Table~\ref{table:llama_guard}. While assessment with \texttt{Llama Guard~3} leads to the same trend (i.e., non-Latin-alphabet languages are less safe than Latin-alphabet languages), it is more conservative with between 4 and 5 times lower ASR values overall compared to \texttt{Claude Sonnet~3.5} numbers. 
Manual review of classification results at the conversation level suggests that \texttt{Llama Guard~3} is missing on a lot of actual unsafe responses that \texttt{Claude Sonnet~3.5} is able to capture.
%(see examples in Appendix~\ref{sec:safety_assessor}).
In practice, the model is tuned for high precision, namely we can trust when it flags a response as unsafe, but is likely to miss less obvious unsafe responses. 
Consequently, reporting ASR numbers with different assessors might give the (false) impression that LLMs are safer than they actually are. 
Again, this highlights the importance of carefully choosing the components of an automation pipeline, as results might not reflect the actual safety risks of a given system.

% * Performance diff across Languages - avg target models    
%     * Bar plot with across languages
% * Performance across turns 
%     * English vs non english
% * Perf across seeds (human, mistral, mixtral, multijail) - all english starting point
%     * turn graph 
% * Perf for starting attack seeds in english vs target language ( multijail in 3 languages)

% \begin{figure}
%     \centering
%     \includegraphics[width=\linewidth]{figures/seed_influence.png}
%     \caption{Caption}
%     \label{fig:seed_influence}
% \end{figure}

% Figure~\ref{fig:seed_influence}

\section{Conclusion}
% \section{Future Work and Limitations}
We present MM-ART, a method for automatically conducting multi-turn and multi-lingual red teaming on black box LLMs. 
From a few conversation starters, our method automatically generates more starters and automatically conduct adversarial conversations against any target LLMs in a wide range of languages.
We showed that multi-lingual LLMs are not uniformly safe across their supported languages and that machine translation can bypass model alignment.
Moreover, the robustness of LLMs with unsafe queries deteriorates with conversation depth.
Through our analysis, we found that translation and multi-turn attacks have compounding effect on the ASR, reaching up to 195\% higher than with standard English single-turn approach.
In the future, we will explore various techniques for regenerating prompts upon LLM refusal \cite{Russinovich2024GreatNW,chainattacksemanticdrivencontextual}.
We plan to reduce even more reliance on human-crafted prompts by leveraging zero-shot generation \cite{2022zeroshot}.
Finally, most recent models support modalities beyond text and we will expand our work to support those.
% are multi-modal, we will expand this line of work to handle deeper conversation and more input modalities.

% With this research, we also encourage more studies and benchmark will be considered mitigation that our paper encourages more curation of multi-lingual safety-related datasets and benchmarks, and improves the overall robustness of safety mechanisms in future LLMs
% \begin{itemize}
%     \item Regeneration of Prompts when face with Refusal

%     \item Utilising a sampling strategy in the seed selection to make the intial set diverse with variety of prompts including innocuous/benign, with jailbreak templates and role play instructions utilising TAP/PAIR/PAP approaches.

%     \item Start from scratch: generate seed prompt from key-words or topics
%     \item we plan to try this with larger model like Mistral Large having more detailed instructions and Policy based exemplars into this component.
%     \item deeper conversation, more than 5 turns
%     \item multi-modal (cf. Llama 3.2 understands images)

% \end{itemize}

\clearpage
\newpage
% \mathrm{https://aclrollingreview.org/cfp#long-papers}
\section{Limitations}
As discussed in the main body of the paper, the choice of safety assessors is important as it determines the safety level of a given target model.
While we manually reviewed examples classified by the different assessors, a more systematic human overwatch should be considered for production pipeline. 
On the conversation starter \texttt{Human} dataset, we only considered a single group of humans for this generation, which might lead to a lack of diversity within the different categories.
Similarly, our in-context learning framework means the generated prompts are tied to human seeds in the context. 
As mitigation, we plan to use our synthetic data generation pipeline to select prompt based on diversity and create more elaborated instructions and more capable models to reduce even further the reliance on human generated seeds. 
We also plan to use framework such as PAIR \cite{PAIR2023} or TAP \cite{tap2024} to rephrase prompts until they reach a certain quality (that we need to define). 
While the results show the strength of our multi-turn approach, we will put more emphasis on the evaluation of the generated turns.
More specifically, we need to evaluate the relevance of the generated turns to the category and the current conversation for a better understanding of the process. 
Similarly, in this work we explored conversations up to 5 turns, and we will explore larger models for automated multi-turn red-teaming that goes way beyond 5 turns, for which the recently released models with large context length are likely to be even more vulnerable.
% \begin{itemize}
    % \item Benchmarking of safety assessors could involve more manual effort
    % \item Diversity: ICL means we can only generate in the area of seeds. Mitigations: regeneration, selection based on diversity, generation from scratch using more capable models and better instructions
    % \item Human-generated prompts : only one group of individuals.
    % \item Multi-turn: more adaptive to combine benign and adversarial, 
    % \item multi-turn: assessment of relevancy of the generated next turn (context-switching is a decent jailbreaker)
    % \item multi-turn: prompt length increases with deeper conversations, so we would have to consider larger models for going deeper.
% \end{itemize}
\section{Ethical Considerations}
Our research investigates the safety challenges of LLMs in multilingual settings. We are aware of the potential misuse of our findings and emphasize that our research is solely for academic purposes and ethical use. Misuse or harm resulting from the information in this paper is strongly discouraged. Through this work, we aim to raise awareness among generative AI practitioners on the risks that come with this technology. We believe identifying and sharing potential risks with the research community is the most efficient path towards mitigation.

% \section{TO ADD}
% \begin{itemize}
%     \item Update more relevant example of conversations
%     \item 3b and 1b results and update all corresponding figures with that
%     \item human vs machine translation
%     \item Approach Diagram[P1]
% \end{itemize}
\clearpage
\newpage

\bibliography{custom}

\appendix
\onecolumn

% \section{Appendix}

\section{Diversity Metrics for Conversation Starters}
\label{app:diversity}
We computed pair-wise cosine distance and SelfBLEU metrics for all 4 conversation starter datasets across every safety category as follows. 
For a given category, we average the pair-wise distance over 10 randomly sampled prompts and repeat the process 100 times to remove the statistical variance among dataset because of their size difference. 
The score of the category is the average of these 100 scores.
We then average across the different safety categories to get the final score for a dataset.
% We used random sampling 100 times to remove the statistical variance among dataset because of their size difference. 
Cosine distance is in similar range (0.74 - 0.79) for machine generated and \texttt{Multi-Jail} datasets while \texttt{Human} set is the most semantically diverse. 
For syntactic comparison using SelfBLEU, we see a similar pattern where \texttt{Multi-Jail} and machine generated datasets are in same range (3.44 - 4.22) while \texttt{Human} set stands at 1.8. We calculated  embedding vector with \texttt{all-MiniLM-L6-v2} model from sentence transformer \cite{sentencetransformer} for calculating Semantic Cosine Distance.
\begin{table}[ht!]
\centering
% \resizebox{\linewidth}{!}{
\begin{tabular}{ccc}
\hline
\textbf{Conversation Starter} & \textbf{Cosine Distance($\uparrow$)} & \textbf{SelfBLEU($\downarrow$)} \\
\hline
Human & 0.84 & 1.8 \\
Multi-Jail & 0.79 & 4.22 \\
\texttt{Mistral-7B} Gen & 0.74 & 3.45 \\
\texttt{Mixtral-8$\times$7B} Gen & 0.77 & 3.44 \\
\hline
\end{tabular}
% }
\captionsetup{justification=centering}
\caption{Diversity Metrics for Conversation Starters}
\label{table:diversity_metrics}
\end{table}

\
\section{Results on Attack Success Rate}

\subsection{Refusal Rate for Conversation Starters}
Here in the Table \ref{table:refusal_rate}, we present the average refusal rate of 1st turn for each conversation starters datasets. This will be used to see the quality of initial prompts.

\begin{table*}[ht!]
\centering
\resizebox{\textwidth}{!}{
\begin{tabular}{cccccccc}
\hline
Dataset & Llama 3.2 - 1B & Llama 3.2 - 3B & Llama 3.1 - 8B & Llama 3.1 - 70b & Mistral Large 2 & Command R+ & Average\\
\hline
\texttt{Human}  &  15.20&  11.60& 23.20 & 9.28 & 5.98 & 4.27 & 11.59\\
\texttt{Multi-Jail} &  45.67&  49.92& 65.49 & 52.22 & 42.70 & 39.00 & 49.17\\
\texttt{Mistral-7B} &  30.82&  28.20& 40.89 & 29.37 & 22.99 & 14.71 & 27.83\\
\texttt{Mixtral-8$\times$7B} &  24.86&  26.32& 45.83 & 28.43 & 21.46 & 16.68 & 27.26\\
\hline
Average &  29.14&  29.01& 43.85 & 29.83 & 23.28 & 18.67 & 28.96\\
\hline
\end{tabular}
}
\captionsetup{justification=centering}
\caption{Average refusal rate at 1st turn across conversations starters \& Target models}
\label{table:refusal_rate}
\end{table*}

\subsection{ASR for all conversation Starters across languages}
Here in the Table \ref{table:asr_depth_starters}, we present the average ASR rate of 1st turn and after 5 turns for each conversation starters datasets against all languages. For e.g, ASR\textsubscript{5} value for Human set shows that on average across all target models 44.7\% of times the conversations lead to generating Unsafe content and there is a 72.6\% gain in ASR going from 1st to 5th turn( 25.9\% to 44.7\%).

\begin{table*}[ht!]
\centering
\captionsetup{justification=centering}
\resizebox{\textwidth}{!}{
\begin{tabular}{c|cc|cc|cc|cc|cc}
\hline
Language & \multicolumn{2}{c}{Human} & \multicolumn{2}{c}{Mulit-Jail} & \multicolumn{2}{c}{Mistral Generated}  & \multicolumn{2}{c}{Mixtral Generated} & \multicolumn{2}{c}{Average} \\

 & ASR\textsubscript{1} & ASR\textsubscript{5} & ASR\textsubscript{1} & ASR\textsubscript{5} & ASR\textsubscript{1} & ASR\textsubscript{5} & ASR\textsubscript{1} & ASR\textsubscript{5} & ASR\textsubscript{1} & ASR\textsubscript{5} \\
English(en) & 25.9 & 44.7 & 17.9 & 33.0 & 17.1 & 36.3 & 33.4 & 49.6 & 23.6 & 40.9 \\
\hline
\multicolumn{11}{c}{Latin Languages}\\
\hline
Spanish(es) & 26.2 & 51.6 & 15.4 & 34.1 & 16.6 & 39.6 & 33.1 & 53.4 & 22.8 & 44.7 \\
French(fr) & 27.2 & 56.1 & 17.0 & 37.5 & 18.5 & 43.9 & 34.7 & 56.4 & 24.4 & 48.5 \\
German(de) & 29.3 & 58.8 & 22.1 & 44.0 & 22.2 & 49.3 & 37.1 & 60.0 & 27.7 & 53.0 \\
Latin(es, fr, de) & 27.6 & 55.5 & 18.2 & 38.5 & 19.1 & 44.3 & 34.9 & 56.6 & 25.0 & 48.7 \\
\hline
\multicolumn{11}{c}{Non-Latin Languages}\\
\hline
Arabic(ar) & 38.7 & 67.0 & 25.5 & 46.7 & 28.6 & 56.0 & 44.2 & 66.5 & 34.3 & 59.0 \\
Hindi(hi) & 41.2 & 72.2 & 33.0 & 57.8 & 32.8 & 61.3 & 46.4 & 72.0 & 38.3 & 65.8 \\
Japanese(ja) & 39.5 & 75.0 & 40.0 & 64.4 & 38.8 & 69.2 & 50.4 & 75.8 & 42.2 & 71.1 \\
average & 32.6 & 60.8 & 24.4 & 45.4 & 24.9 & 50.8 & 39.9 & 61.9 & 30.5 & 54.7 \\
Non-Latin(hi, ja, ar) & 40.3 & 73.6 & 36.5 & 61.1 & 35.8 & 65.2 & 48.4 & 73.9 & 40.3 & 68.5 \\

\hline
\end{tabular}
}
\caption{Attack Success Rate (ASR) after 1 turn (ASR\textsubscript{1}) and 5 turns (ASR\textsubscript{5}) for each conversation starter dataset}
\label{table:asr_depth_starters}
\end{table*}

\vspace{12\baselineskip}

\subsection{ASR for all target models across languages}
Here in the Table \ref{table:asr_depth_compare_target}, we present the average ASR rate of 1st turn and after 5 turns for each target model against all languages. This table is similar to the Table\ref{table:asr_lang} present here in main paper but with values of ASR\textsubscript{1}. It helps us to see the improvement in ASR for each tagret model and languages combination in going from 1st to 5th turn.

\begin{table*}[ht!]
\centering
\captionsetup{justification=centering}
\resizebox{\textwidth}{!}{
\begin{tabular}{c|cc|cc|cc|cc|cc|cc|cc}
\hline
Language & \multicolumn{2}{c}{Llama 3.2 - 1B} & \multicolumn{2}{c}{Llama 3.2 - 3B} & \multicolumn{2}{c}{Llama 3.1 - 8B} & \multicolumn{2}{c}{Llama 3.1 - 70b} & \multicolumn{2}{c}{Mistral Large 2} & \multicolumn{2}{c}{Command R+} & \multicolumn{2}{c}{Average} \\

 & ASR\textsubscript{1} & ASR\textsubscript{5} & ASR\textsubscript{1} & ASR\textsubscript{5} & ASR\textsubscript{1} & ASR\textsubscript{5} & ASR\textsubscript{1} & ASR\textsubscript{5} & ASR\textsubscript{1} & ASR\textsubscript{5} & ASR\textsubscript{1} & ASR\textsubscript{5} & ASR\textsubscript{1} & ASR\textsubscript{5} \\
\hline
English(en) & 31.4 & 58.8 & 24.6 & 40.5 & 13.0 & 27.0 & 23.5 & 41.5 & 24.5&  40.2& 24.4& 37.6& 23.6& 40.9 \\
\hline
\multicolumn{15}{c}{Latin Languages}\\
\hline
Spanish(es) & 30.4 & 64.5 & 22.6 & 50.4 & 12.8 & 29.3 & 18.3 & 38.0 & 19.3& 37.7 & 33.8& 48.2 & 22.8& 44.7 \\
French(fr) & 31.5 & 68.3 & 25.9 & 61.4 & 11.5 & 31.4 & 20.4 & 37.4 & 21.2& 41.9 & 35.6& 50.3 & 24.3& 48.5 \\
German(de) & 44.7 & 80.1 & 30.4 & 64.4 & 13.7 & 34.6 & 21.3 & 41.6 & 22.3& 45.2 & 33.7& 52.3 & 27.7& 53.0 \\

Latin - Average & 35.5 & 70.9 & 26.3 & 58.7 & 12.6 & 31.8 & 20.0 & 39.0 & 20.9& 41.6 & 34.4& 50.2 & 25.0& 48.7 \\
\hline
\multicolumn{15}{c}{Non-Latin Languages}\\
\hline
Arabic(ar) & 46.1 & 74.6 & 41.6 & 71.2 & 20.7 & 45.1 & 27.0 & 47.3 & 57.5 & 57.5 & 58.4 & 58.4 & 34.3& 59.0 \\
Hindi(hi) & 53.6 & 87.6 & 50.4 & 80.9 & 26.5 & 51.5 & 29.9 & 54.9 & 56.9 & 56.9 & 63.1 & 63.1 & 38.3& 65.8 \\
Japanese(ja) & 59.4 & 94.2 & 54.4 & 84.9 & 35.0 & 65.4 & 36.5 & 62.9 & 58.2 & 58.2 & 60.9 & 60.9 & 42.2& 71.1 \\
Non-Latin Average & 56.5 & 90.9 & 52.4 & 82.9 & 30.8 & 58.4 & 33.2 & 58.9 & 29.4& 57.6 & 39.4& 62.0 & 40.3& 68.5 \\
\hline
All - Average & 42.4 & 75.4 & 35.7 & 64.8 & 19.0 & 40.6 & 25.2 & 46.2 & 25.2& 48.2 & 35.1& 53.0 & 30.5& 54.7 \\
\hline

\end{tabular}
}
\caption{Attack Success Rate (ASR) after 1 turn (ASR\textsubscript{1}) and 5 turns (ASR\textsubscript{5}) for all Target Models}
\label{table:asr_depth_compare_target}
\end{table*}

% \vspace{12\baselineskip}

\subsection{ASR after Refusal response in Initial Prompt}
\label{sec:asr_after_refusal}
Here in the Table \ref{table:asr_after_refusal}, we present the average ASR rate after 5 turns for the conversations where initial prompt leads to Refusal response. Across all conversations( target models and languages), the average ASR is 54.7\%, the average refusal rate of the first response is 29\%(refer to \ref{table:refusal_rate} and on those 29\% conversations, the ASR drops to 6.64\%. This justifies the claim that from an initial refusal, it is harder to lead the conversation to a successful attack, as the refusal remains in the context until the end of the conversation. 

\begin{table*}[ht!]
\centering
\captionsetup{justification=centering}
\resizebox{\textwidth}{!}{
\begin{tabular}{cccccccc}
\hline
Language & Llama 3.2 - 1B & Llama 3.2 - 3B & Llama 3.1 - 8B & Llama 3.1 - 70b & Mistral Large 2 & Command R+ & Average\\
\hline
English(en) & 5.76& 2.65& 5.07 & 4.37 & 1.92 & 3.21 & 3.83\\
% \hline
\hline
\multicolumn{8}{c}{Latin Languages}\\
\hline
Spanish(es) & 13.62& 9.89& 7.02 & 5.03 & 2.82 & 2.32 & 6.78\\
French(fr) & 18.84& 11.87& 8.73 & 6.07 & 5.50 & 4.00 & 9.17\\
German(de) & 14.08& 10.38& 9.76 & 5.06 & 3.15 & 3.46 & 7.65\\
Latin - Average & 15.51& 10.71& 8.50 & 5.38 & 3.82 & 3.26 & 7.87\\
\hline
\multicolumn{8}{c}{Non-Latin Languages}\\
\hline
Arabic(ar) & 11.88& 11.33& 9.60 & 3.30 & 3.43 & 2.06 & 6.94\\
Hindi(hi) & 11.74& 6.65& 6.31 & 3.83 & 7.72 & 4.07 & 6.72\\
Japanese(ja) & 8.31& 5.68& 10.62 & 3.79 & 3.28 & 2.30 & 5.66\\
Non-Latin Average & 10.03& 6.17& 8.84 & 3.64 & 4.81 & 2.81 & 6.19\\
\hline
All - Average & 12.06& 7.85& 8.16 & 4.49 & 3.97 & 3.06 & 6.64\\
\hline

\end{tabular}
}
\caption{Attack Success Rate (ASR) for conversations where the initial response is a ``refusal``.}
\label{table:asr_after_refusal}
\end{table*}

% \vspace{5\baselineskip}
\vspace{13\baselineskip}

\section{Refusal Assessor Benchmarking}
\label{app:refusal}

\begin{table}[ht!]
\centering
% \resizebox{\linewidth}{!}{
\begin{tabular}{ccccc}
\hline
\textbf{name} & \textbf{precision} & \textbf{recall} & \textbf{f1} & \textbf{f0\.5} \\
\hline
Rule-based & 0.95 & 0.78 & 0.85 & 0.91 \\
\texttt{Mistral-7B} & 0.93 & 0.98 & 0.95 & 0.94 \\
\texttt{Mixtral-8$\times$7B} & 0.99 & 0.84 & 0.91 & 0.95 \\
\hline
\end{tabular}
% }
\captionsetup{justification=centering}
\caption{Refusal Benchmarking}
\label{table:refusal_benchmark}
\end{table}

\section{Multi-Jail Mapping Categories}
Here, we present the mapping of Multi-Jail category that we used to map with categories of our MM-ART Framework.

\begin{table}[ht!]
\centering
% \resizebox{\linewidth}{!}{
\begin{tabular}{cc}
\hline
\textbf{Multi-jail category} & \textbf{Our category} \\
\hline
Animal abuse & Animal Abuse \\Self-harm& Self Injury and Harmful Dieting \\
Adult content & Sexual Content \\
Sexual exploitation \& human trafficking & Sexual Content \\
Discrimination \& injustice & Inclusivity \\
misinformation regarding ethics, laws and safety & Harmful Misinformation \\
Conspiracy theories \& misinformation & Harmful Misinformation \\
Soliciting personally identifiable information & Private Information and Personal Details \\
Substance abuse \& banned substances & Dangerous Devices and Substances \\
Weapons & Dangerous Devices and Substances \\
\hline
\end{tabular}
% }

\captionsetup{justification=centering}
\caption{Mapping of Multi-Jail categories with our categories}
\label{table:multijail_mapping}
\end{table}

\section{Hyperparameters}
For inference with target models, we used max new tokens as 1000 and temperature of 0.5 with top\_p of 0.9. Rest all parameters are kept default. For attack prompts generation with \texttt{Mixtral 8x7B}, we used max new tokens as 512, temperature as 0.9 and top\_p as 0.9.

\end{document}